\documentclass[letterpaper, 10 pt, conference]{ieeeconf}  

\IEEEoverridecommandlockouts                              

\usepackage{graphicx}
\usepackage{subfigure}
\usepackage{caption}
\usepackage{hyperref} 
\usepackage{microtype}
\usepackage{makecell}
\usepackage{amsmath}
\usepackage{amsfonts}
\usepackage{bm}
\usepackage{color}
\usepackage{multirow}
\usepackage[ruled,linesnumbered]{algorithm2e}
\SetAlFnt{\small}
\SetAlCapFnt{\small}
\SetAlCapNameFnt{\small}
\usepackage{etoolbox}
\usepackage{cite}
\usepackage{threeparttable}
\usepackage{booktabs}
\apptocmd{\thebibliography}{\raggedright}{}{}
\usepackage{geometry}

\title{\LARGE \bf
CL-MAPF: Multi-Agent Path Finding for Car-Like Robots \\  with Kinematic and Spatiotemporal Constraints 
}

\author{Licheng~Wen$^{1}$, Zhen~Zhang$^{1}$, Zhe~Chen$^{1}$, Xiangrui~Zhao$^{1}$, and Yong~Liu$^{1,*}$
\thanks {$^{1}$Licheng~Wen, Zhen~Zhang, Zhe~Chen, Xiangrui~Zhao and Yong Liu are with the State Key Laboratory of Industrial Control Technology and Institute of Cyber-Systems and Control, Zhejiang University, Zhejiang, 310027, China (*Yong Liu is the corresponding author, {\tt\small yongliu@iipc.zju.edu.cn})}
}

\begin{document}

\maketitle
\thispagestyle{empty}
\pagestyle{empty}

\begin{abstract}
    Multi-Agent Path Finding has been widely studied in the past few years due to its broad application in the field of robotics and AI. However, previous solvers rely on several simplifying assumptions. They limit their applicability in numerous real-world domains that adopt nonholonomic car-like agents rather than holonomic ones. In this paper, we give a mathematical formalization of Multi-Agent Path Finding for Car-Like robots (CL-MAPF) problem. For the first time, we propose a novel hierarchical search-based solver called Car-like Conflict-Based Search to address this problem. It applies a body conflict tree to address collisions considering shapes of the agents. We introduce a new algorithm called Spatiotemporal Hybrid-State A* as the single-agent path planner to generate path satisfying both kinematic and spatiotemporal constraints. We also present a sequential planning version of our method for the sake of efficiency. 
    We compare our method with two baseline algorithms on a dedicated benchmark containing 3000 instances and validate it in real-world scenarios. The experiment results give clear evidence that our algorithm scales well to a large number of agents and is able to produce solutions that can be directly applied to car-like robots in the real world.
    The benchmark and source code are released in \url{https://github.com/APRIL-ZJU/CL-CBS}.
\end{abstract}

\section{Introduction}

Multi-Agent Path Finding, also known as MAPF, is a crucial planning problem for multiple agents. Each agent is required to move from an initial start place to a specified goal and avoid collisions with each other. Due to its broad applications in AI and robotics community, research on MAPF has been flourishing in the past few years.

MAPF is known to be an NP-hard problem\cite{yu2013structure}. Famed approaches to solve this problem can be classified into reduction-based methods\cite{yu2013planning,surynek2015reduced,erdem2013general}, A*-based methods\cite{wagner2011m,standley2010finding,ferner2013odrm,phillips2011sipp}, prioritized methods\cite{ji2007computational}\cite{vcap2015prioritized} and dedicated search-based methods\cite{sharon2015conflict,boyarski2015icbs,felner2017search}. Some researches also take partial kinematic constraints into consideration\cite{honig2016multi,li2019multi,ma2019lifelong,yakovlev2019prioritized}.

MAPF can be applied to several contemporary scenarios including self-driving cars,
autonomous straddle carriers\cite{dobrev2017steady}, warehouse robots\cite{sartoretti2019primal}, unmanned surface vehicles\cite{wen2020} and office robots\cite{veloso2015cobots}. These industrial and service robots are generally nonholonomic and designed as car-like vehicles in practice.
However, almost all the above methods base on assumptions that agents are modelled as disks and are capable of rotating in place. These solvers also adopt discrete 4-neighbor grids as their search space.

In reality, car-like robots are in nature with rectangle shapes and have minimum turning radii.
Original MAPF solvers can be applied by reducing the grid-graph resolution and adopting dedicated controllers to track generated paths. However, this may generate coarser solutions and degrade their practical applicability since the controllers cannot track paths precisely, especially those with sharp turns.
These solvers also apply various types of conflicts, including vertex conflicts and edge conflicts, to avoid collisions between moving agents\cite{stern2019multi}.
Nevertheless, the types of conflicts adopted in different situations depend on their specific environments, and they can not represent all the collision scenarios.

To address these concerns, it is essential to formalize \textit{Multi-Agent Path Finding for Car-Like robots} (CL-MAPF) problem.
We propose a novel hierarchical search-based solver called Car-Like Conflict-Based Search(CL-CBS) to settle this problem.

Our main contributions are summarized as follows:
\begin{itemize}
    \item We present a complete CL-MAPF solver, which simply uses body conflicts to describe all inter-agent collision scenarios. Our approach also ensures the robustness for agents' execution error.
    \item We propose a new single-agent path planner for car-like robots, which generates path satisfying both kinematic and spatiotemporal constraints.
    \item We also introduce a sequential version of our original method, which significantly reduces search time with little sacrifice on performance.
    \item We conduct experiments in both simulated and physical environment. They demonstrate our method can scale well to large amount of agents and produce solutions directly applied to car-like robots in real-world scenes.

\end{itemize}

\begin{figure}[tb]
    \centering
    \subfigure[WeTech Robot]{
        \includegraphics[width=0.24\textwidth]{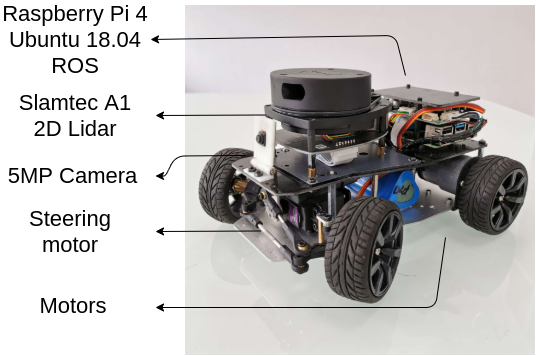}
        \label{nanocar}
    }
    \subfigure[Snapshot of the field test]{
        \includegraphics[width=0.198\textwidth]{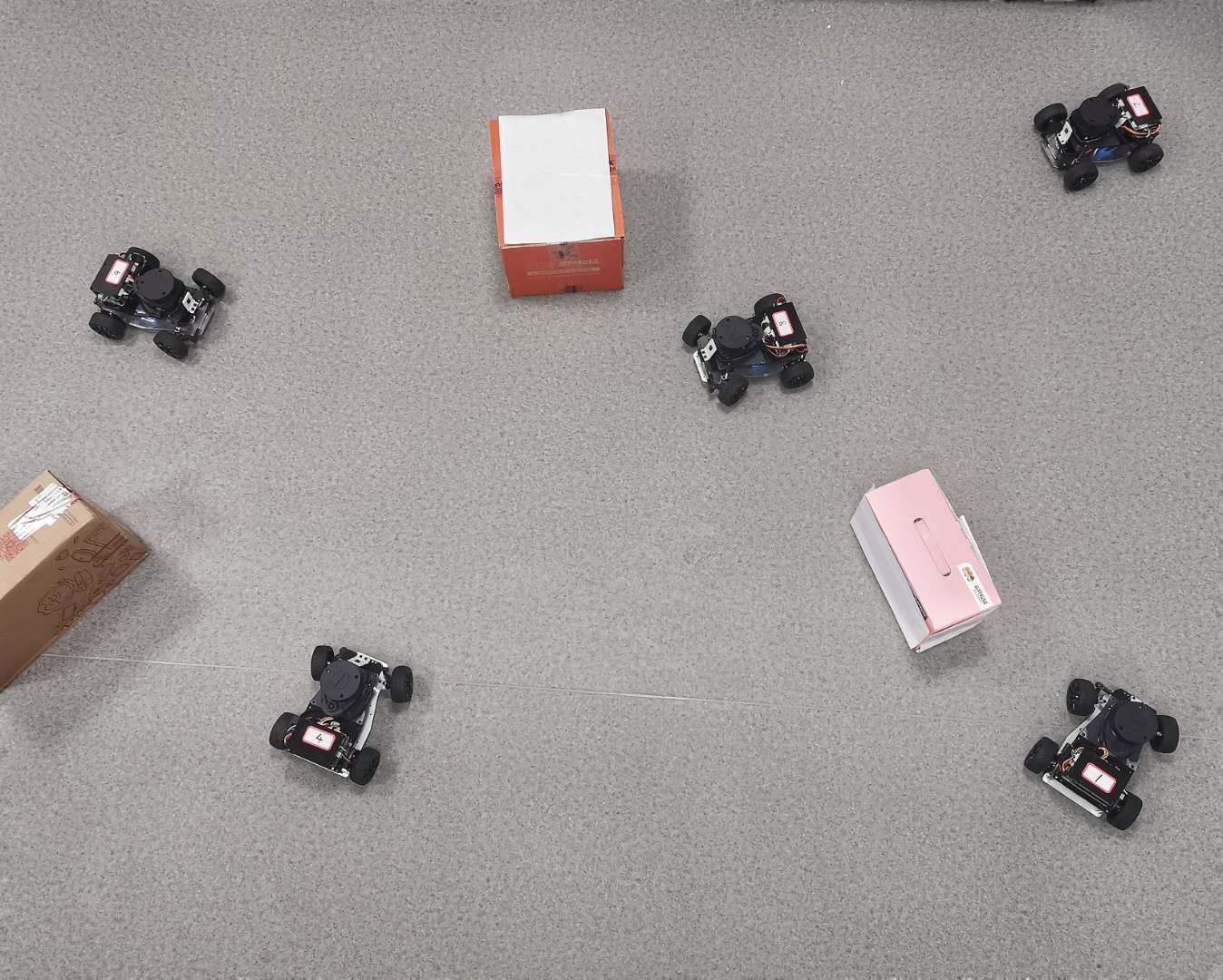}
        \label{experiment}
    }
    \caption{Our proposed method tested on seven ackermann-steering robots produced by WeTech. The experiment video is available in the attachment of this paper.}
    \label{fig1}
\end{figure}

\section{Related works}
\label{Related works}

MAPF problem has been widely studied in the robotic and AI community. One way to solve the problem is  reduction to other well-studied combinatorial problems\cite{yu2013planning,surynek2015reduced,erdem2013general}. Besides, several solvers using search techniques have been proposed to solve this problem.
M*\cite{wagner2011m} expands search nodes to all possibilities when conflict occurs. OD-recursive-M* (ODrM*)\cite{ferner2013odrm} adapts the concept of Operator Decomposition\cite{standley2010finding} to keep the branching factor small. Another complete and optimal MAPF solver is the Safe Interval Path Planning (SIPP)\cite{phillips2011sipp}. It runs an A* search in a graph where each node represents a pair of vertexes in the workspace and a safe time interval.
One popular branch of MAPF solvers nowadays is based on a two-level optimal solver called  Conflict-Based Search(CBS)\cite{sharon2015conflict}. ICBS\cite{boyarski2015icbs} and CBSH\cite{felner2017search} improves CBS further by classifying conflicts and resolving cardinal conflicts first.
Finally, a prioritized approach\cite{vcap2015prioritized} is also a common choice in numerous cases. However, it lacks a completeness guarantee.

Most of the methods above use some assumptions, like ignoring robot's kinematic constraints and using discrete grid graphs.
MAPF-POST algorithm that works on differential-drive robots is proposed in \cite{honig2016multi}. It takes velocity limits into account and provides a guaranteed safety distance between robots. \cite{li2019multi} presents a generalized version of CBS for large agents that occupy more than one grid.
SIPPwRT\cite{ma2019lifelong} combines the token passing algorithm with SIPP for pickup and delivery scenarios.
In \cite{yakovlev2019prioritized} a road-map based planner supporting different moving speeds is suggested, and a grid-based planner capable of handling any-angle moves using a variant of SIPP is proposed in \cite{yakovlev2017any}.

There are also researches about distributed collision avoidance for multiple nonholonomic robots. Traditional approaches for single robot can be applied, including artificial potential field\cite{khatib1986real}, dynamic window approach\cite{brock1999high}, and model predictive control\cite{morgan2014model}. The reciprocal velocity obstacle (RVO)\cite{van2008reciprocal} is a decentralized algorithm allowing robots to avoid each other with no communication between them.
Optimal reciprocal collision-avoidance (ORCA)\cite{van2011reciprocal} succeeds the concept of velocity obstacle and solves the problem faster by casting into a low-dimensional linear program.  Bicycle reciprocal collision avoidance (B-ORCA)\cite{alonso2012reciprocal} and $\epsilon$CCA\cite{alonso2018cooperative} are two adaptions of ORCA for car-like vehicles.
Nevertheless, ORCA-based methods need global path planners to avoid deadlocks in scenarios with obstacles and cannot guarantee that each robot can reach its goal.

\section{CL-MAPF Problem}
\label{problem def}
Classic MAPF solvers usually consider holonomic agents moving in cardinal directions and neglect agents' size. This will cause the generated solutions cannot be executed on real-world multi-agent systems, especially for those composed of car-like robots.
In this section, we first present the robot kinematic model and then present the definition of Multi-Agent Path Finding for Car-Like robot (CL-MAPF) problem.

\begin{figure}[tb]
    \centering
    \begin{minipage}[t]{0.5\linewidth}
        \centering
        \centerline{\includegraphics[width=\textwidth]{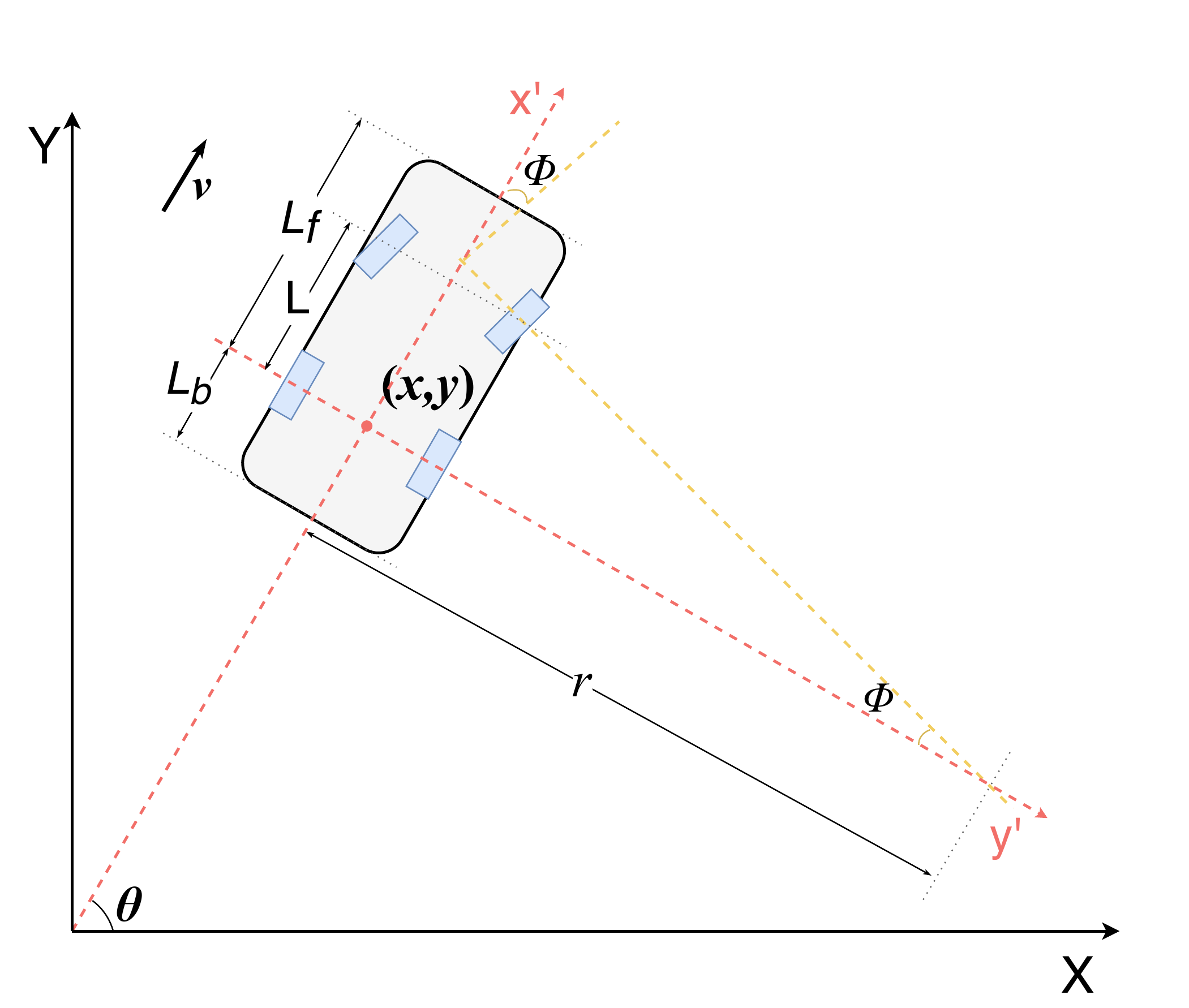}}
        \label{ackermann model}
    \end{minipage}
    \begin{minipage}[t]{0.32\linewidth}
        \centering
        \includegraphics[width=\textwidth]{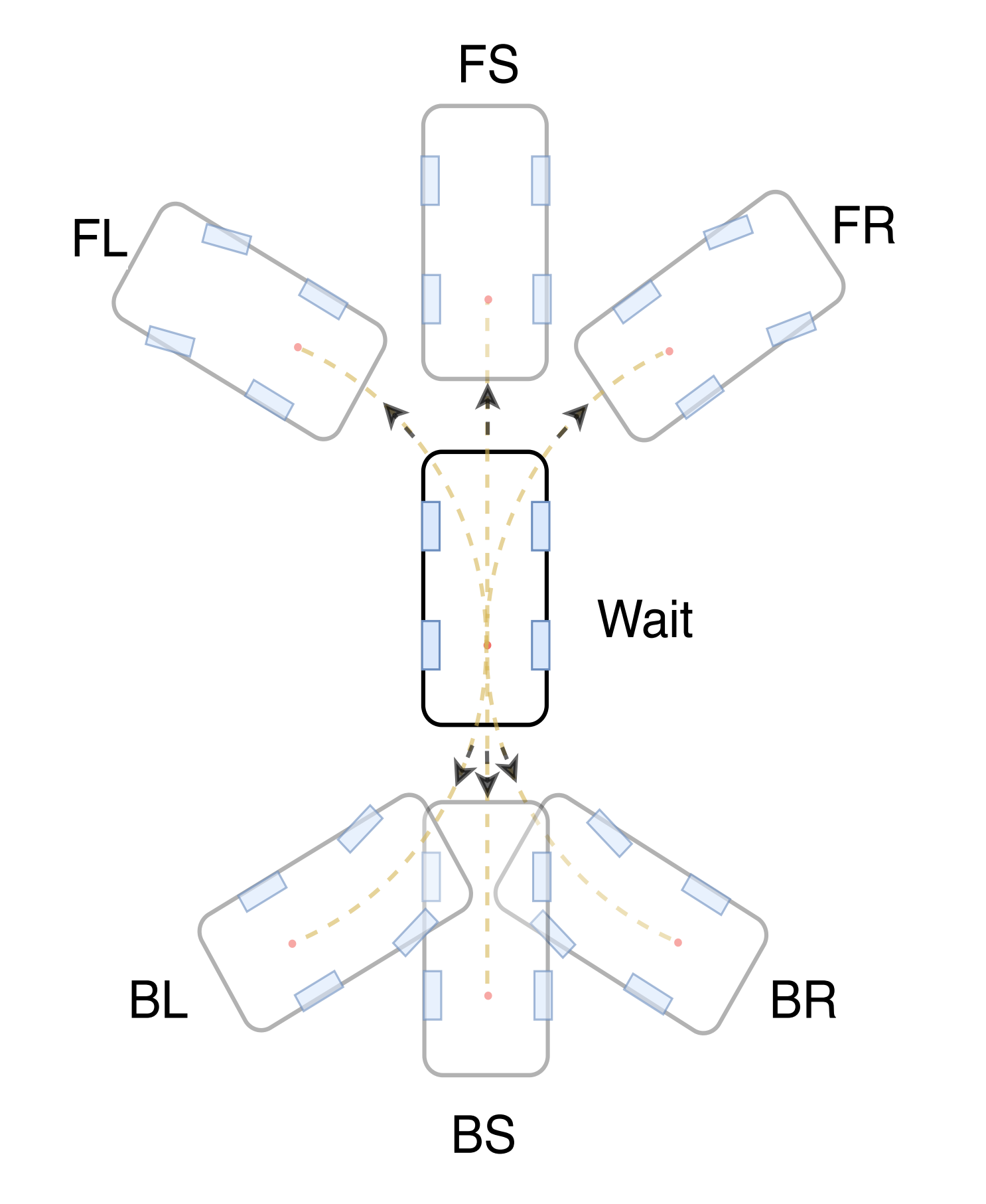}
        \label{ackermann steering}
    \end{minipage}
    \caption{Ackermann-steering model
    }
    \label{ackermann}
\end{figure}

\begin{figure*}[t]
    \centering
    \includegraphics[width=0.95\textwidth]{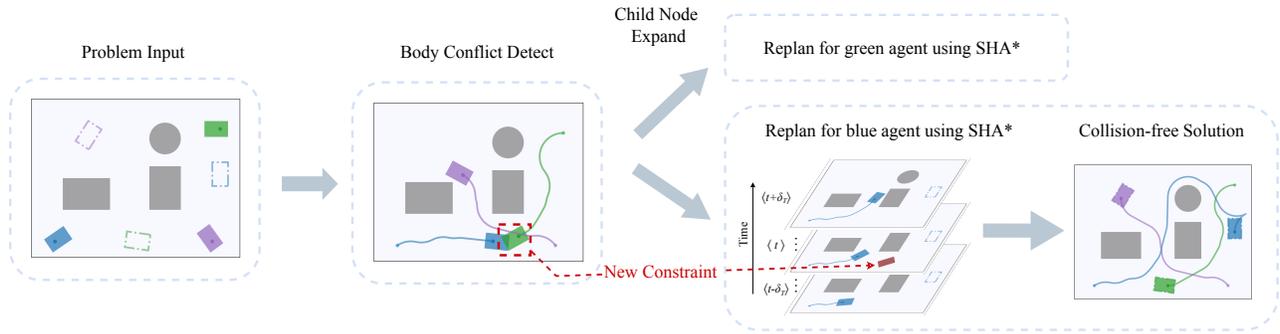}
    \caption{
        A pipeline of Car-like CBS. Agents' start states are denoted as solid colored rectangle and goal states as dotted outline rectangles. Grey area represents the obstacle region $\mathbf{O}_{ws}$. A body conflict between blue agent and green agent is detected in middle figure. Then two child nodes are expanded with each contains a new constraint and spatiotemporal hybrid-state A* is performed for the agent receiving it. }
    \label{pipeline}
    \vspace{-15pt}
\end{figure*}

\subsection{Robot Kinematic Model}
\label{kinematic_model}

Kinematic constraints must be considered for nonholonomic robots. Several path models like circular trajectories, asymptotically heading trajectories apply to different kinds of robots in practice. For car-like robots discussed in this paper, we commonly use Ackermann steering geometry as the kinematic model shown in Fig.\ref{ackermann}. The kinematic constraints forbid it to move laterally and rotate in place \cite{marin2013event}.

The state for an Ackermann-steering robot is denoted by $\mathbf{s} = (x,y,\theta)$. The origin of rigid body frame $(x,y)$ places at the center of robot's rear axle. The x-axis of body frame points alongside yaw angle $\theta$, y-axis points to the left side of the robot. $v$ represents the robot's velocity, and $\phi$ represents the steering angle of front wheels. When the steering angle is fixed at $\phi$, radius of the circular trajectory which robot moves along is denoted as $r = L/\tan\phi$. Let $dw = r\cdot{d\theta}$ represents the distance along trajectory after time $dt$.

The kinematic relation between $\phi$ and $\dot{\theta}$ is defined as:
\begin{equation}
    \dot{\theta}  =  \frac{v}{L}\tan\phi
\end{equation}

By discretizing and recursively integrating with sample time $T_s$ , we can calculate robot state at timestep $t$ as following:
\begin{equation}
    \mathbf{s}_t =
    \left[
        \begin{array}{c}
            x      \\
            y      \\
            \theta \\
        \end{array}
        \right]_{t}
    =
    \left[
        \begin{array}{c}
            x      \\
            y      \\
            \theta \\
        \end{array}
        \right]_{t-1}
    +T_{s}
    \left[
        \begin{array}{c}
            v \cos \theta \\
            v \sin \theta \\
            \frac{v}{L}\tan\phi
        \end{array}
        \right]_{t-1}
    \label{ackermann-equation}
\end{equation}

The robot's velocity $v$ is bounded as $v_{bmax}\leq v \leq v_{fmax}$, where $v_{bmax} < 0$ and $v_{fmax} > 0$ represent the max speed when robot moves forward or backward, respectively. The steering angle is restricted by $\phi_{max}$, which implies each Ackermann-steering robot should maintain a minimum turning radius $r_{min}$ during the whole path.

\subsection{Problem Definition}

Given a workspace $\mathbf{W} \subset \mathbb{R}^3 $ and a set of obstacles occupying an arbitrary region $\mathbf{O}_{ws} \subset \mathbf{W}$, we formalize CL-MAPF problem as follows.

There are $K$ car-like agents $a_1,a_2,...,a_K$. Time is assumed to be discretized. Let $\mathbf{s}_t^i$ be the state of agent $a_i$ at timestep $t$. The start and goal state of $a_i$ is respectively denoted as $\mathbf{s}_{start}^i \in \mathbf{W} $ and $\mathbf{s}_{goal}^i \in \mathbf{W}$. A \textit{single-agent path} $p^i = [\mathbf{s}_0^i,...,\mathbf{s}_{T_i}^i,\mathbf{s}_{T_i+1}^i,...]$ for $a_i$ is \textit{feasible} iff all the following four conditions are satisfied:

\begin{itemize}
    \item [a.]
          Path of $a_i$ should begin at its start state and stops at its goal states after limited timesteps $T_i$. That is $p^i[0] = \mathbf{s}_{start}^i$ and $p^i[T_i]= \mathbf{s}_{goal}^i$.
    \item [b.]
          Agent $a_i$ would stay at the goal position after reaching it,  $\forall t\geq T_i$, $p^i[t]= \mathbf{s}_{goal}^i$.
    \item[c.]
          Agent $a_i$ should never collide with obstacles at any timestep $t$.
    \item [d.]
          Each move of an Ackermann-steering agent should satisfy the kinematic model. Thus given agent's max forwarding speed $v_{fmax}$, max reversing speed $v_{rmax}$ and  max steering angle $\phi_{max}$, agent state $p^i[t]$ and $p^i[t+1]$ should obey Equation \ref{ackermann-equation} for any timestep $t$.
\end{itemize}

As shown in Fig.\ref{ackermann}, we use $L_f$ and $L_b$ to denote distance from rear axle to robot front and robot rear, respectively. $W_r$ denotes the width of robot. For an agent at state $(x_0,y_0,\theta_0)$, the rectangle shape of agent body  $C$ in Cartesian coordinate system can be defined as:
\begin{equation}
    \begin{array}{c}
        C=\left\{(x, y) \in \mathbb{R}^{2}, f(x, y)\leq 2\right\} \\
        f(x, y) = \left|\frac{x'}{L_b+L_f}+\frac{y'}{W_r}\right|+\left|\frac{x'}{L_b+L_f}-\frac{y'}{W_r}\right|
    \end{array}
    \label{rectangle_overlap}
\end{equation}
where,
\begin{equation*}
    \left[
        \begin{array}{c}
            x' \\
            y' \\
        \end{array}
        \right] =
    \left[\begin{array}{cc}
            \cos \theta_0  & \sin \theta_0 \\
            -\sin \theta_0 & \cos \theta_0
        \end{array}\right]
    \left[
        \begin{array}{c}
            x-x_0 \\
            y-y_0 \\
        \end{array}
        \right]+
    \left[
        \begin{array}{c}
            L_b           \\
            \frac{W_r}{2} \\
        \end{array}
        \right]
\end{equation*}

We use a tuple  $\langle a_{i},  a_{j}, C_{t}^i, C_{t}^j, t\rangle$ to denote a \textit{collision} , or a \textit{body conflict}, between agent $a_i$ and agent $a_j$ at timestep $t$ when $ C_{t}^i \cap C_{t}^j \neq \emptyset$. $C_{t}^i$ represents body rectangle for $a_i$ at timestep $t$.

A \textit{solution} for CL-MAPF problem is a set of feasible paths for all K agents where every two paths are \textit{collision-free}. A CL-MAPF example is shown in problem input of Fig.\ref{pipeline}.

Solutions can be evaluated using two commonly used functions: makespan and sum of costs. Since agents are not required to move at the same speed each timestep, definitions of these two functions slightly vary from classic MAPF.  Makespan is the maximum length of paths for all agents to reach their goals, $\max_{i \in [1,K]} d(p^i)$. Sum of costs is the total length of all agents' paths, $\sum_{i=1}^{K} d(p^i)$.
Since the original MAPF problem is proven to be NP-hard\cite{yu2013planning}, finding an optimal solution for CL-MAPF are also NP-hard.

\section{Methodology}
\label{Methodology}

We introduce a novel solver for CL-MAPF problem called Car-like CBS. The high-level body conflict search tree is a variant of the conflict tree in CBS. As for low-level path finding method for single agent, we proposed Spatiotemporal Hybrid-State A* algorithm to cope with both kinematic and spatiotemporal constraints. We also introduce a sequential planning version of our method at the end of this section. It remarkably shortens the searching time with little sacrifices on completeness.

\subsection{Body Conflict Tree}
\label{Inter-agent Constraint Tree}

The classic MAPF solvers apply various types of conflicts (the most common ones are vertex conflicts and edge conflicts) to avoid collisions between two single-agent paths. Yet these conflicts cannot represent all situations of agent colliding. Benefit from planning in a continuous workspace, we can simply use body conflicts to describe all inter-agent collision scenarios. We propose a binary \textit{body conflict tree} (BCT) and perform best-first search on it. Each node on BCT contains a set of inter-agent constraints and a solution that satisfies these constraints.

The expansion of the BCT works is shown in Fig.\ref{pipeline}.
When the leaf node $N$ with minimum cost popped out from BCT, a collision check is executed for the solution belonging to $N$. If a body conflict $\langle a_{i},  a_{j}, C_{t}^i, C_{t}^j, t\rangle$ has been detected, we produce two inter-agent constraints: $\langle a_{i}, C_{t}^j,[t-\delta_T,t+\delta_T] \rangle$ for $a_i$ and $\langle a_{j}, C_{t}^i,[t-\delta_T,t+\delta_T] \rangle$ for $a_j$. The former constraint denotes that agent $a_i$ should not pass through rectangle area $C_{t}^j$ from timestep $t-\delta_T$ to time step $t+\delta_T$, likewise the latter. Then two child node of $N$ are generated, each contains one of inter-agent constraints.
At last we perform a low-level search in both of child nodes for the agent received the extra constraint.
The pseudocode of BCT is shown from line \ref{alg:1} to line \ref{alg:2} in Algorithm \ref{algorithm}.

In the view of robots will not execute as we expect in practice, our method retains certain robustness in both time and space dimensions.  When there are execution errors on position for agents, we inflate the rectangle area $C_t$ of inter-agent constraint. By multiplying an inflation coefficient  $k$ on param $L_f$,  $L_b$ and $W_c$ in Equation \ref{rectangle_overlap}, robots possess bigger space for others to bypass. The param $\delta_T$ of the constraint definition is used for eliminating errors in the time dimension.

\begin{algorithm}[tbp]
    \SetAlgoSkip{}
    \caption{Car-like CBS}
    \label{algorithm}
    \SetKwFunction{FSum}{SH\_Astar}
    \SetKwProg{Fn}{Function}{:}{}
    \SetKw{Continue}{continue}

    $Root.constraints\leftarrow \emptyset$\;\label{alg:1}
    $Root.plan\leftarrow$ path for each agent using $SH\_Astar(a_i)$\;
    BCT$\leftarrow\{Root\}$\;
    \While{BCT$\neq \emptyset$}
    {
        $Node \leftarrow \arg \min _{N^{\prime} \in \operatorname{BCT}} N^{\prime}. \operatorname{cost}$\;
        BCT $\leftarrow$ BCT $\backslash\{Node\}$\;
        $C\leftarrow CollisionDetect(Node.plan)$\;
        \If{C=$\emptyset$}
        {
            \KwRet $Node.plan$\;
        }
        \ForEach{$C_i = \langle a_{i}, C_{t}^j,[t-\delta_T,t+\delta_T] \rangle \in C$}{
            $New.plan\leftarrow Node.plan$\;
            $New.constraints\leftarrow Node.constraints\cup \{C_i\}$\;
            $New.plan\  for\ a_i \leftarrow SH\_Astar(a_i)$\;
            \If{$SH\_Astar(a_i)\neq \emptyset$}
            {
                BCT $\leftarrow$ BCT $\cup\left \{New\right\}$\;
            }
        }
    }\label{alg:2}
    \Fn{\FSum{$a_i$}}{\label{alg:3}
    $Open\leftarrow \{ (0,x_{start}^i,y_{start}^i,\theta_{start}^i)\}$\;
    \While{$Open\neq \emptyset$}
    {
        $N \leftarrow \arg \min _{N^{\prime} \in \operatorname{Open}} N^{\prime}. \operatorname{fScore}$\;
        \If{$(N.x,N.y,N.\theta)$ near $\mathbf{s}_{goal}^i$}
        {
            $path_{toGoal} \leftarrow AnalyticExpand(\mathbf{s}_{goal}^i)$\;
            \If{$ CollisionDetect(path_{toGoal}) = \emptyset$}
            {
                \KwRet $path_{whole}$\;
            }
        }
        \ForEach{act $\in$ steering actions}{
            $N^{\prime}=(N.t+T_s, x, y, \theta)\leftarrow Expand(N,act)$\;
            \If{\textbf{not} SatisfyConstraint($N^{\prime}$) \textbf{or}  $N^{\prime} \in \mathbf{O}_{ws} $ }
            {\Continue}
            $N'.gScore= N.gScore+ penalty(act)\times cost(act)$\;
            $N'.h= N'.gScore+ heuristic(N',\mathbf{s}_{goal}^i)$\;
            \uIf{$N'\notin Open$}
            {
                $Open \leftarrow Open \cup\left \{N'\right\}$\;
            }
            \ElseIf{$N'.gScore < N_{inOpen}.gScore$}
            {
                update Node with $N'.state, N'.gScore$
            }
        }
    }
    \KwRet $\emptyset$\;
    }\label{alg:4}
\end{algorithm}

\subsection{Spatiotemporal Hybrid-State A*}
\label{Spatiotemporal Hybrid-State A*}

As mentioned above, the high-level body conflict tree requires low-level search to:
\begin{itemize}
    \item Plan paths satisfying the kinematic constraint to be executed by Ackermann-steering agents;
    \item Plan paths satisfying spatiotemporal inter-agent constraints with other agents;
    \item Discretize paths by sample time $T_s$ and return a state sequence $p[t]$;
\end{itemize}

A well-known path planner applied to the continuous 3D state space for car-like robots is Hybrid-State A* \cite{dolgov2008practical}, but it cannot deal with spatiotemporal constraints. We proposed an adaptation called Spatiotemporal Hybrid-State A* (SHA*) as the low-level planner for Car-like CBS.

When planning for multiple agents, the ability to stay still at the current state in order to avoid other agents moving is necessitated. Thus, the seven steering actions for expanding child nodes are forward max-left(FL), forward straight(FS), forward max-right(FR), backward max-left(BL), backward straight(BS), backward max-right(BR), and wait, as shown in Fig.\ref{ackermann}. We denote actions besides moving straight and waiting as \textit{turning actions}.

Spatiotemporal hybrid-state A* uses a 4D search space $(t, x, y, \theta)$, with $x,y,\theta\in \mathbb{R}$ and time $t$ being discrete. For a node with search state $(t_0, x_0, y_0, \theta_0)$, its child nodes will have states like $(t_0+T_s, x_1, y_1, \theta_1)$, where $(x_1, y_1, \theta_1)$ are computed using robot kinematic Equation \ref{ackermann-equation}.
When a node popped from the open list, we use seven different steering actions to expand this node. For each of seven child states, we not only check collisions with obstacles in $\mathbf{O}_{ws}$, but also check if the  state satisfies all the spatiotemporal constraints for this agent.
Spatiotemporal hybrid-state A* is indicated from line \ref{alg:3} to line \ref{alg:4} in Algorithm \ref{algorithm} as \textit{SH\_Astar} function.

It is worth noted that we add three penalties to cost function $gSocre$ when the agent  performs turning actions, driving backwards, and switching the moving direction. The heuristic function design and analytic expansion technique of our method are the same as the original hybrid-state A*.

\begin{figure}[tb]
    \centering
    \subfigure[Sequential car-like CBS]{
        \begin{minipage}[t]{0.20\textwidth}
            \centering
            \includegraphics[width=\textwidth]{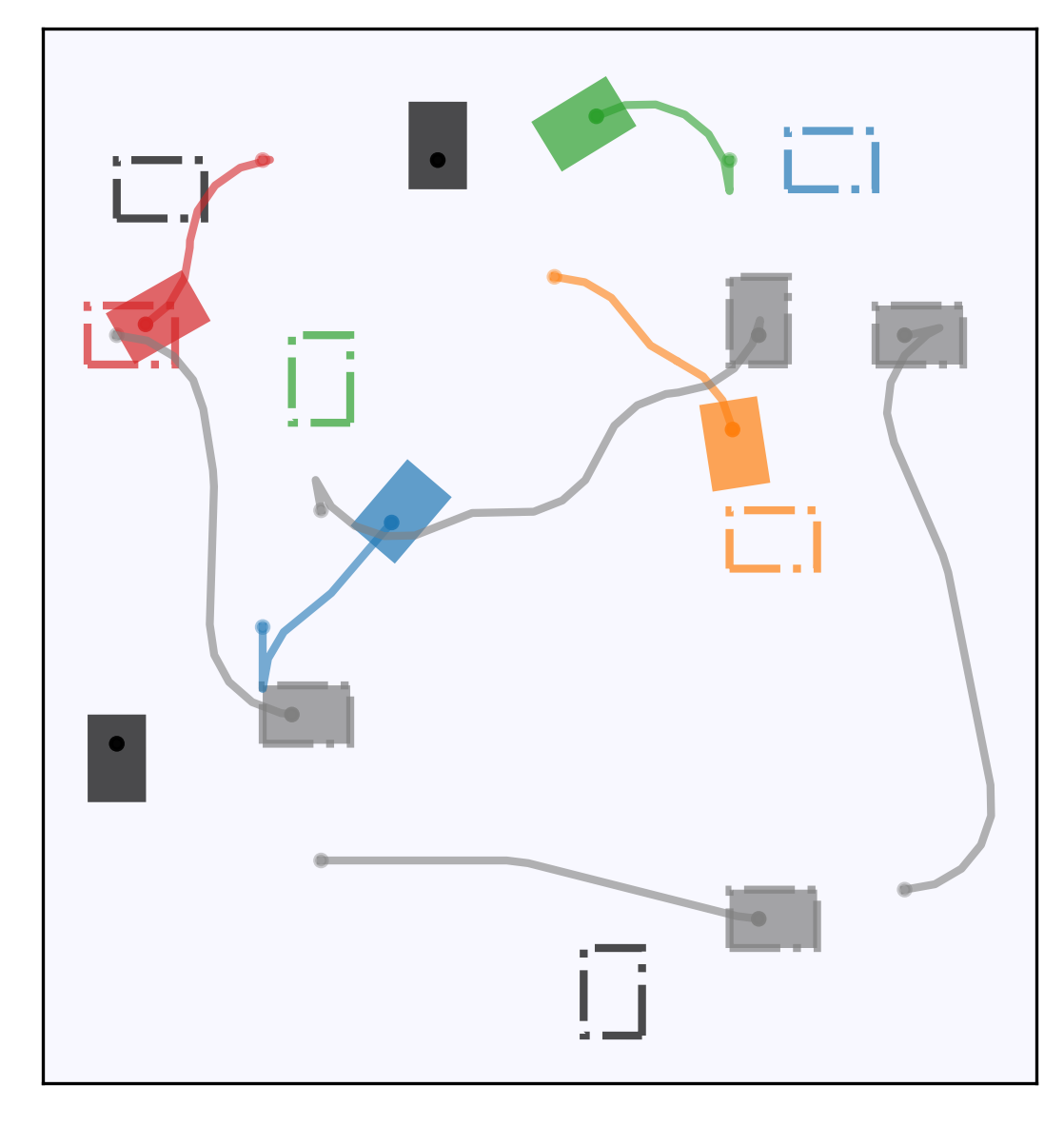}
            \label{sequential version}
        \end{minipage}}
    \subfigure[A fail case]{
        \begin{minipage}[t]{0.1625\textwidth}
            \centering
            \includegraphics[width=\textwidth]{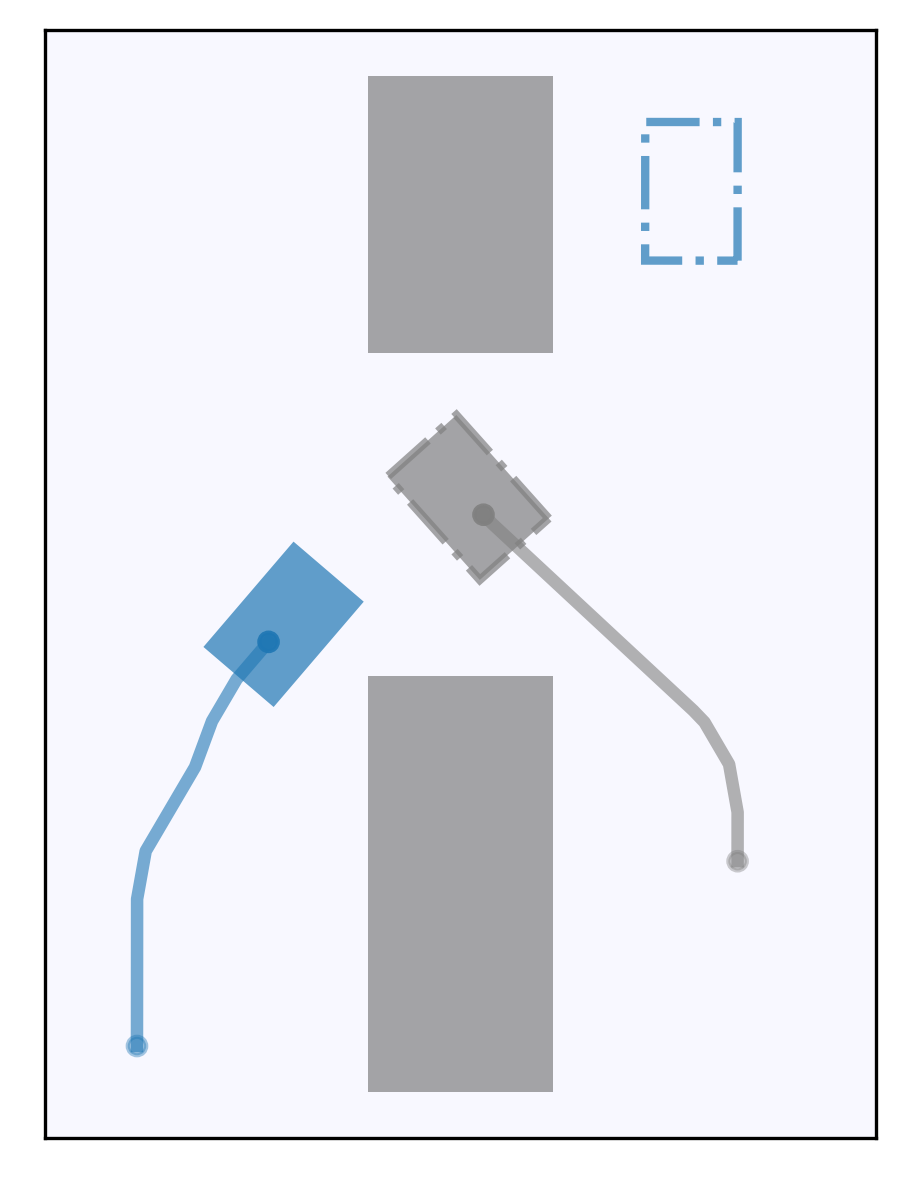}
            \label{sequential fail}
        \end{minipage}}
    \caption{(a) Sequential CL-CBS. (b) A simple fail case for sequential CL-CBS. The blue agent cannot reach its goal when the grey agent planned in former batch arrives at its goal (which locates between the obstacles)  before the blue one passing through those obstacles. }
\end{figure}

\begin{figure*}[tb]
    \centering
    \includegraphics[width=0.9\textwidth]{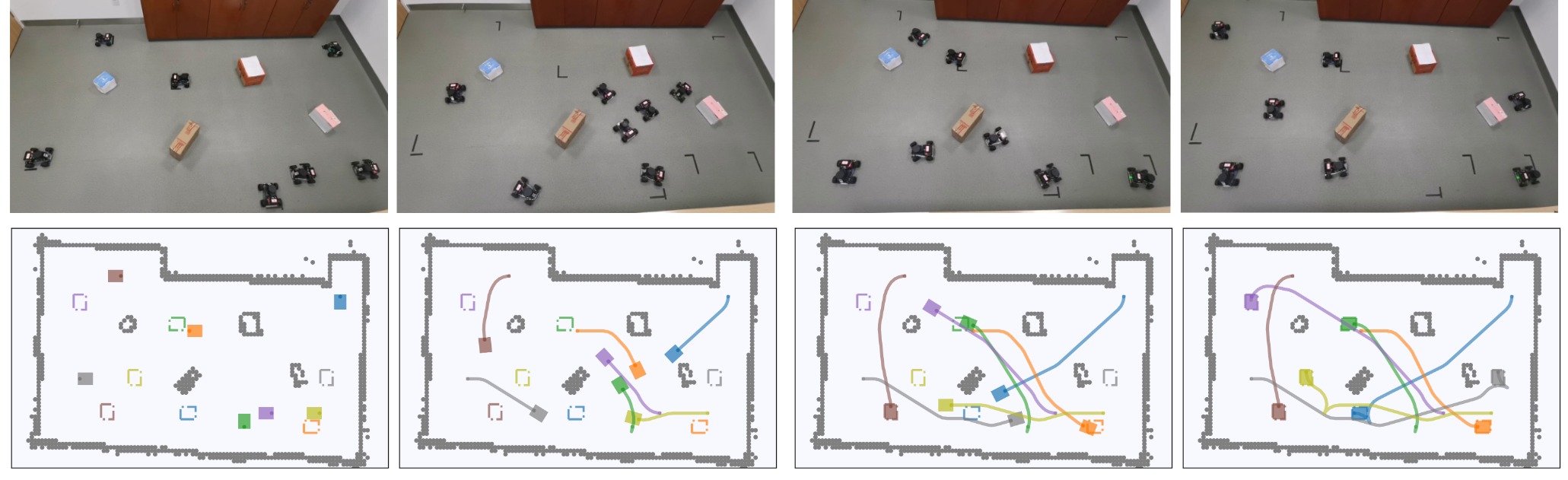}
    \caption{Glimpses of field tests. Snapshots in the upper row show four keyframes during an experiment, and pictures in the lower row plot the trajectories agents have driven at the corresponding frame. Best viewed in color. }
    \label{experiment snapshot}
    \vspace{-15pt}
\end{figure*}

\subsection{Sequential Car-like CBS}
\label{Sequential Car-like CBS}

In car-like CBS solver, the high-level conflict search tree is proven to be both optimal and complete\cite{ma2019searching}. However, spatiotemporal Hybrid-State A* search only ensures completeness but lacks theoretical optimality guarantees.
Thus, the whole car-like CBS solver is complete and near-optimal.

As a result of expanding workspace from discrete space to continuous space, the low-level search time suffers from scalability problems when the number of obstacles increasing and the workspace getting larger. Besides, the high-level search tree expands more nodes when multiple agents visiting the same region at the approximately same time.
These will lead to a noticeable increase in the searching time of Car-like CBS.

Though the scalability problem of the single-agent planner is unavoidable, we propose a sequential planning method to reduce high-level search time inspired by \cite{park2020efficient}.
We divide the $K$ agents into $K_b$ batches, and each batch contains $\lceil K/K_b \rceil$ agents except the last batch. Then we sequentially solve these sub-CL-MAPF problems for each batch and combines result paths together as the final solution of the whole problem. For a batch $b$, the actions of agents in subsequent batches are ignored. The paths planned out in former batches act as dynamic obstacles in the workspace, and they are added to the constraint set of the root node in BCT.

The procedure of this method is exhibited in Fig.\ref{sequential version}.
The agents are divided into three batches. The paths in grey planned out in the first batch act as dynamic obstacles for agents planning in the second batch (colored). Black agents denote agents of the third batch. As a result of avoiding solver to deal with overmuch agents at the same time, the sequential method shortens searching time by nearly an order of magnitude in our experiment. However, it should also be noted that this sequential method sacrifices the completeness guarantee of Car-like CBS. A simple fail case is shown and explained in Fig.\ref{sequential fail}.

\section{Experiments}
\label{Experiments}
In this section, we implement Car-like CBS solver in C++ using \textit{boost} library for math calculation and OMPL library to produce Reeds-Shepp paths. The program are executed on a PC running Ubuntu 16.04 with Intel i7-8700@3.20GHz and 8G RAM. We test our car-like CBS solver in both simulated and physical environment and the experiment video is available in the attachment of this paper.

\smallskip
\subsection{Simulated Experiment}

Since it is the first time we proposed CL-MAPF problem, there are few direct competitors for our experiments. We adopt two methods, a centralized one and a decentralized one, acting as the baseline of our experiment. \textit{i)}  The centralized method is model predictive control with CBS (CBS-MPC) based on \cite{negenborn2009multi}. It applies original CBS solver to provide guide paths for each agent and use MPC to generate final trajectories.  \textit{ii)} The decentralized baseline we use is plain hybrid-state A* (HA*) for a single agent, without the high level-search tree and spatiotemporal constraints.


\subsubsection{Benchmark}
The classic MAPF benchmark like DAO map sets are all 4-neighbor grid-based benchmark and cannot be used for CL-MAPF problem. Therefore we generate a novel CL-MAPF benchmark for simulation experiment.

The benchmark contains two scenarios involving workspace with and without obstacles. Each scenario includes 25 map sets. These map sets possess three type of map size (300$\times$300m,100$\times$100m,50$\times$50m) and distinct agent numbers from 5 to 100. Every map set has 60 unique instances, and the whole benchmark contains 3000 different instances.

For each instance in the benchmark, \textit{i)} it describes a continuous $\mathbb{R}^3$ workspace; \textit{ii)} the start and goal states of agents are guaranteed not collide with each other (for agents under 5$\times$5m size); \textit{iii)} the Euclidean distance between start and goal state of an agent is greater than $1/4$ of the map width; \textit{iv)} for instances with obstacles, it contains circle obstacles with 1m radius and the entire obstacle region occupies 1\% map area.
We use \textit{300x300\_agents80\_obs} to denote mapset with 80 agents in a 300$\times$300m workspace with obstacles.

Based on the benchmark, we evaluate the performance of our method compared with two baseline algorithm HA* and CBS-MPC.

\begin{table}[htbp]
    \centering
    \caption{Comparision with CBS-MPC}
    \resizebox{0.9\linewidth}{!}{
        \begin{tabular}{ccclc}
            \toprule
            \multirow{2}{*}{Mapsize(m$^2$)}              & \multirow{2}{*}{Agents}                 & \multirow{2}{*}{Method} & \multicolumn{2}{c}{Empty / Obstacles}                   \\ \cmidrule(lr){4-5}
                                                         &                                         &                         & Makespan(m)                           & Collision Times \\ \midrule
            \multicolumn{1}{c}{\multirow{2}{*}{300x300}} & \multicolumn{1}{c}{\multirow{2}{*}{50}} & CBS-MPC                 & 206.6/205.4                           & 9.25/10.09      \\
            \multicolumn{1}{c}{}                         & \multicolumn{1}{c}{}                    & Ours                    & \textbf{179.1/178.8}                  & \textbf{0/0}    \\
            \multicolumn{1}{c}{\multirow{2}{*}{100x100}} & \multicolumn{1}{c}{\multirow{2}{*}{30}} & CBS-MPC                 & 71.92/67.90                           & 9.43/9.96       \\
            \multicolumn{1}{c}{}                         & \multicolumn{1}{c}{}                    & Ours                    & \textbf{70.73/67.25 }                 & \textbf{0/0}    \\
            \multicolumn{1}{c}{\multirow{2}{*}{50x50}}   & \multicolumn{1}{c}{\multirow{2}{*}{20}} & CBS-MPC                 & 36.38/35.54*  \tnote{1}               & 7.28/9.26       \\
            \multicolumn{1}{c}{}                         & \multicolumn{1}{c}{}                    & Ours                    & 48.80/52.96                           & \textbf{0/0}    \\ \bottomrule
        \end{tabular}}
    \begin{tablenotes}
        \item[1] *Though having a smaller makespan, we don't consider CBS-MPC performs better due to the collisions in the solution.
    \end{tablenotes}
    \label{table}
    \vspace{-10pt}
\end{table}

\begin{figure*}[tb]
    \centering
    \subfigure[Success rate on 300x300 mapset]{
        \begin{minipage}[t]{0.23\textwidth}
            \centering
            \includegraphics[width=\textwidth]{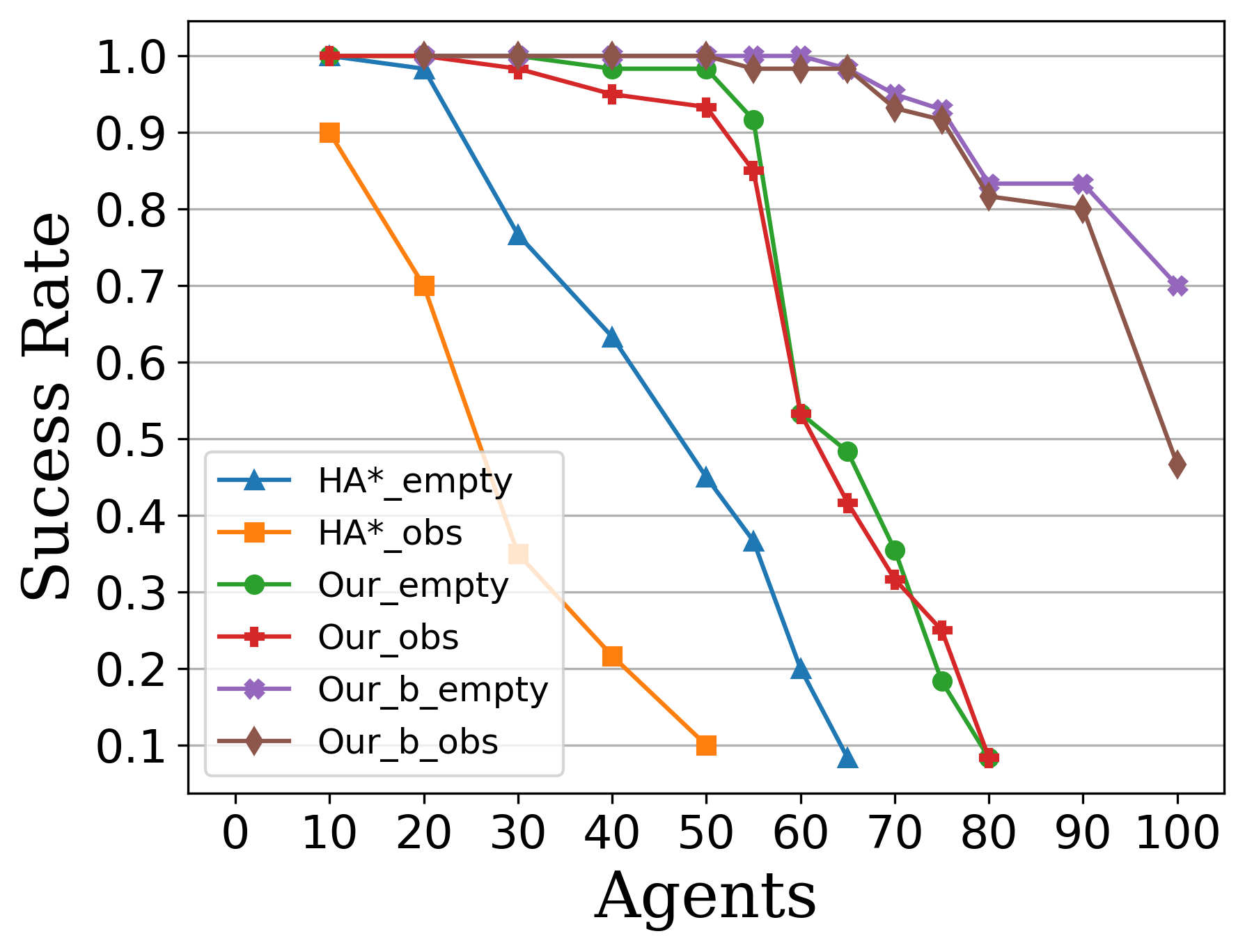}
        \end{minipage}}
    \subfigure[Runtime on 300x300 mapset]{
        \begin{minipage}[t]{0.23\textwidth}
            \centering
            \includegraphics[width=\textwidth]{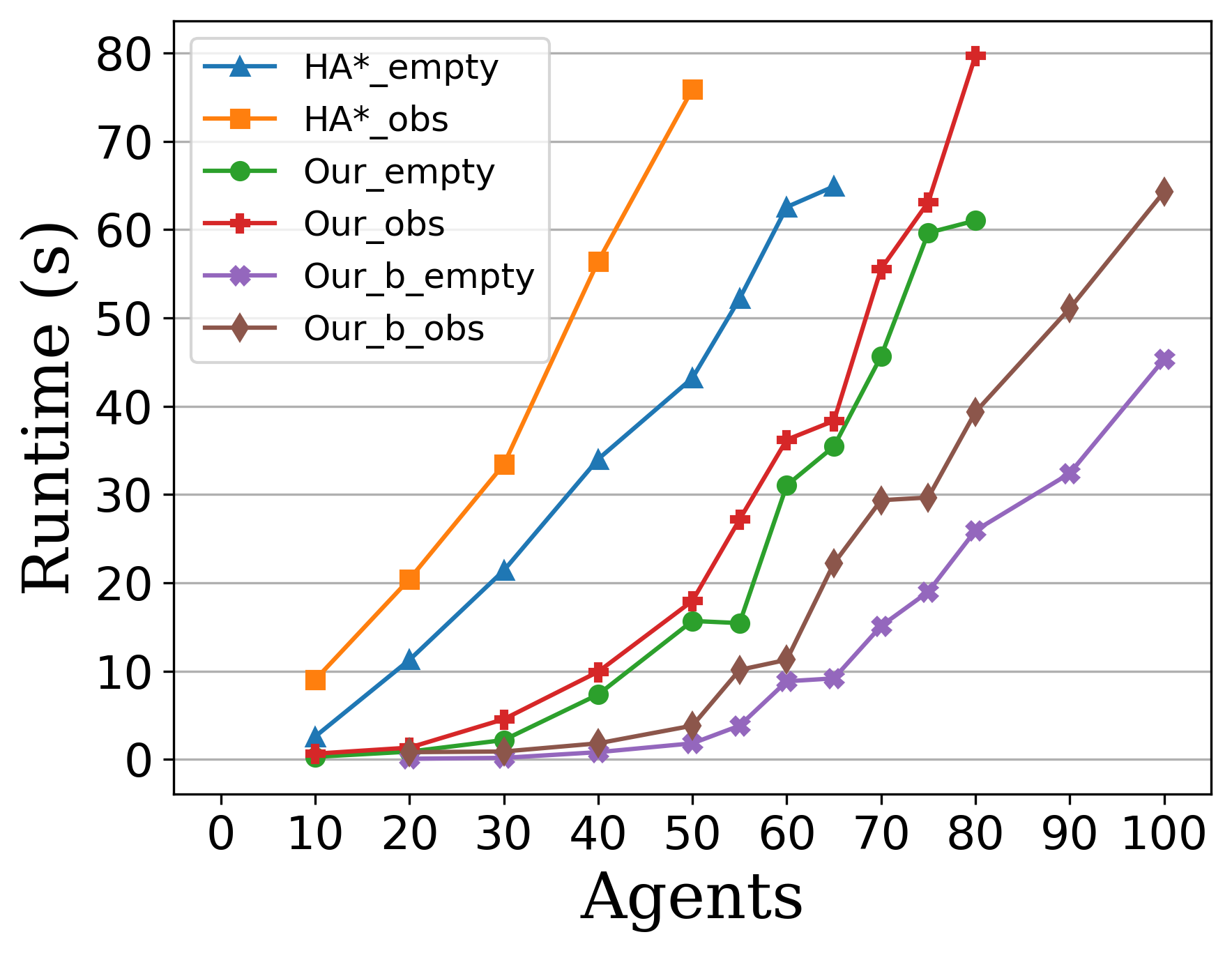}
        \end{minipage}}
    \subfigure[Success rate on 100x100 mapset]{
        \begin{minipage}[t]{0.23\textwidth}
            \centering
            \includegraphics[width=\textwidth]{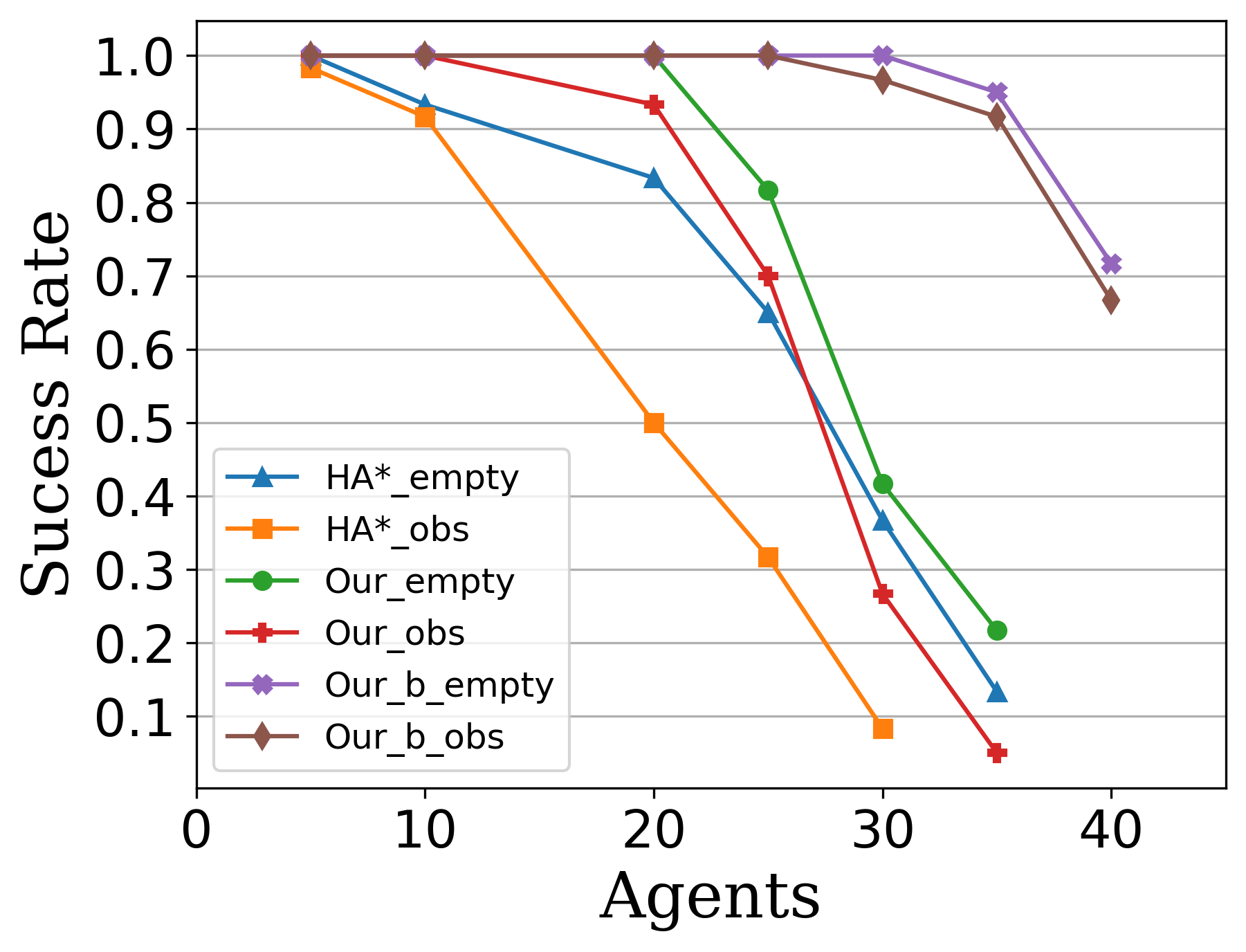}
        \end{minipage}}
    \subfigure[Runtime on 100x100 mapset]{
        \begin{minipage}[t]{0.23\textwidth}
            \centering
            \includegraphics[width=\textwidth]{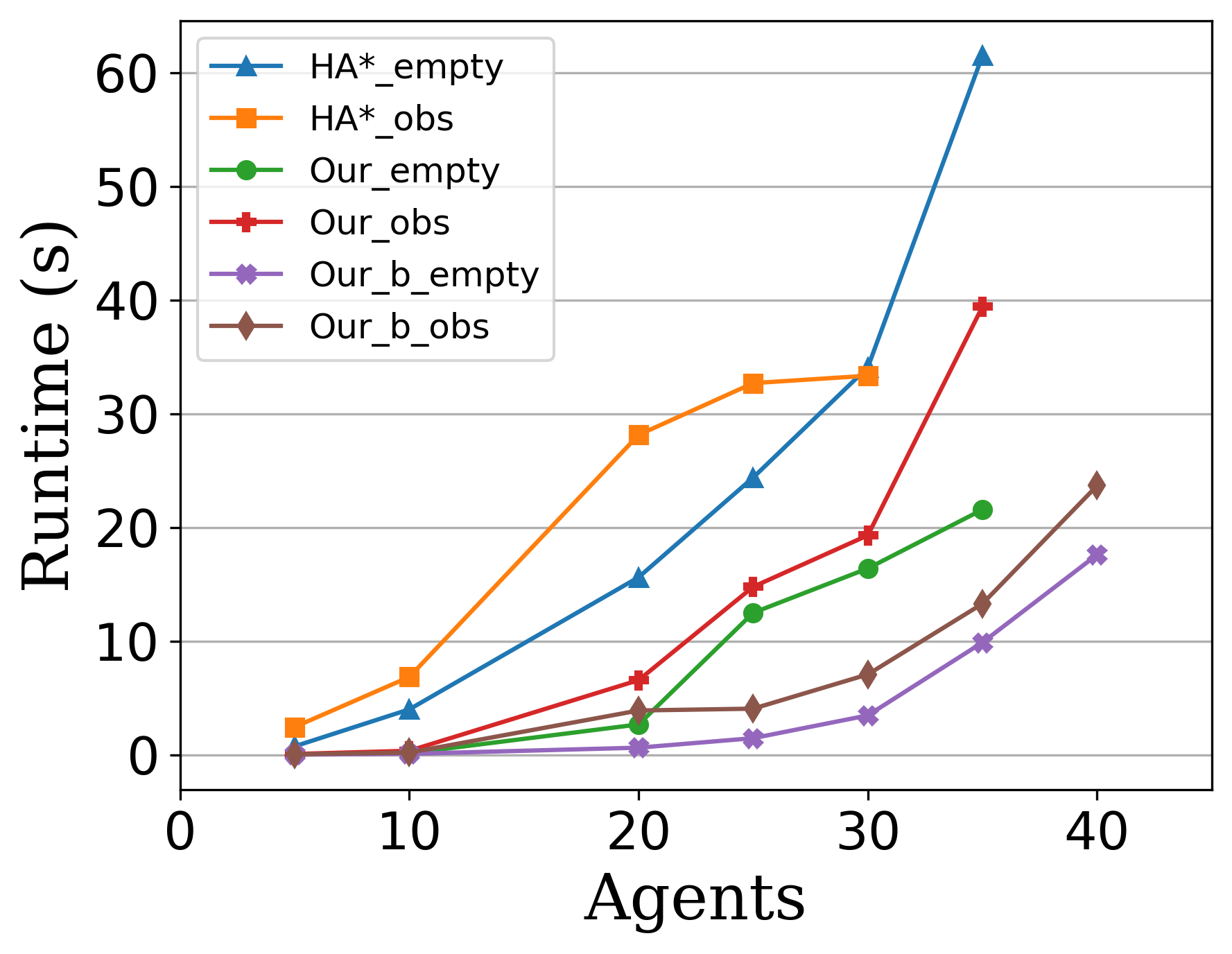}
        \end{minipage}}
    \caption{Scalability Comparision with HA*}
    \label{Scalability result}
    \vspace{-15pt}
\end{figure*}

\begin{figure}[htbp]
    \centering
    \subfigure[Results on 300x300\_agents60\_obs mapset]{
        \begin{minipage}[t]{0.85\linewidth}
            \centering
            \includegraphics[width=\textwidth]{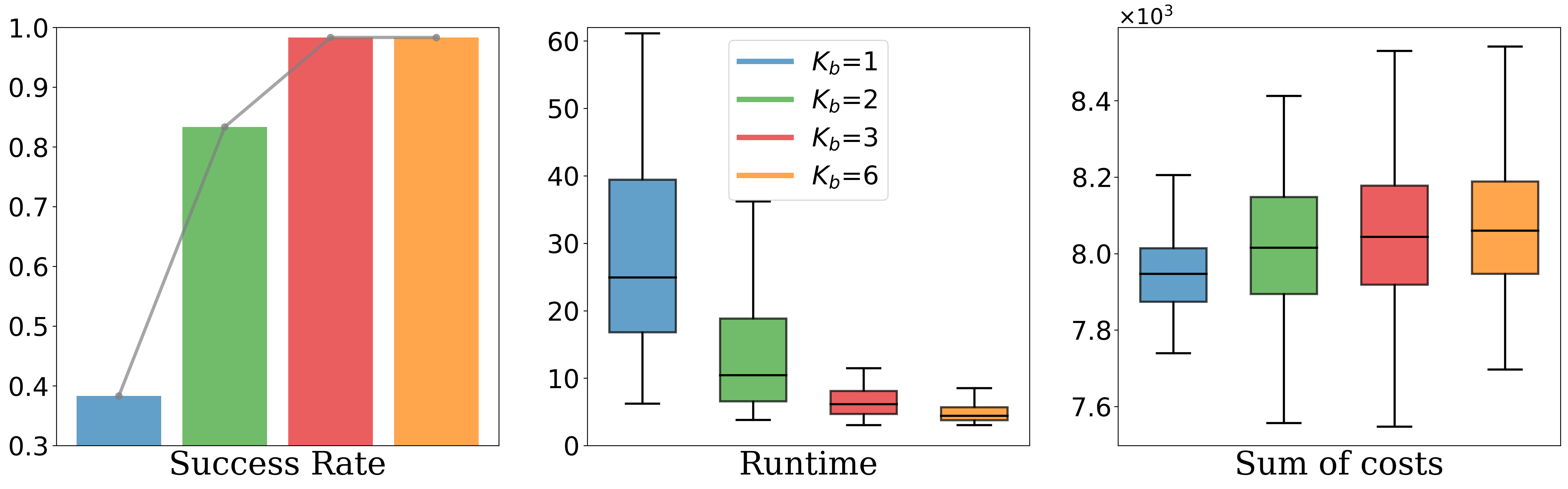}
            \label{Sequential1}
        \end{minipage}}
    \subfigure[Results on 50x50\_agents15\_empty mapset]{
        \begin{minipage}[t]{0.85\linewidth}
            \centering
            \includegraphics[width=\textwidth]{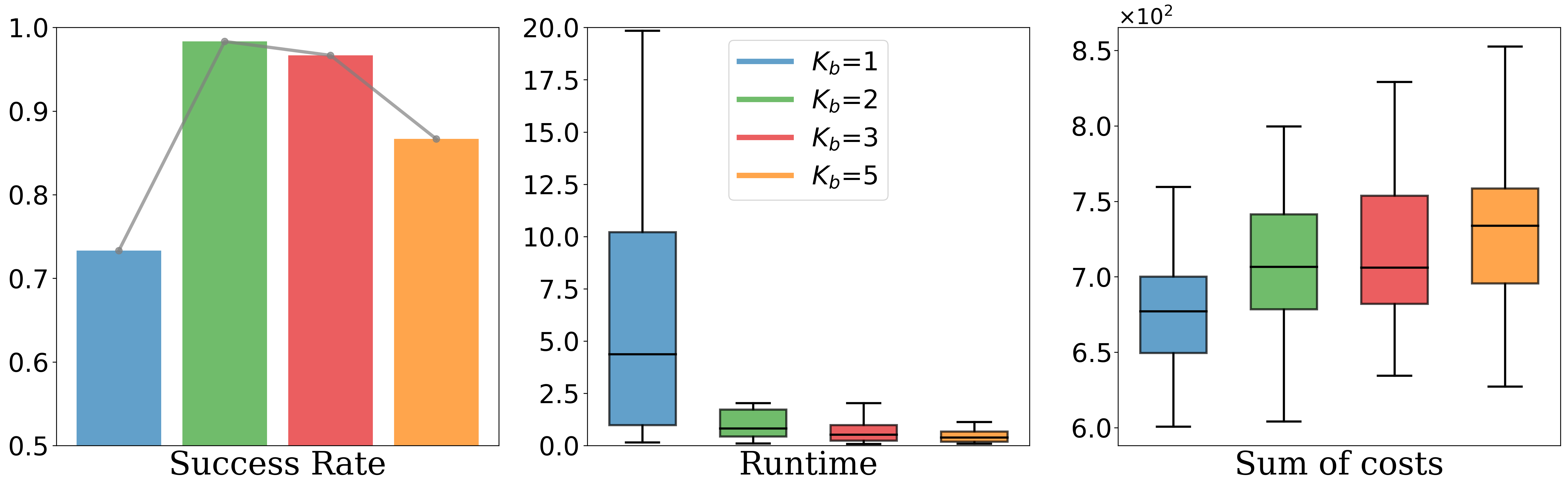}
            \label{Sequential2}
        \end{minipage}}
    \caption{Sequential CL-CBS experiment results}
    \label{Sequential result}
\end{figure}

\subsubsection{Comparision with CBS-MPC}

We assume agents in the experiments are homogeneous with the following parameters: the shape of agents is  2$\times$3m as $L_f=2m, L_b=1m$, the maximum speed for both forward and backwards $v_{max}=2m/s$,  the minimum turning radius $r=3m$. We set the runtime limit for each instance as 90 seconds and compare the average makespan and collision times in the solution of each map set. The results are shown in Table \ref{table}.

Our method outperforms CBS-MPC in almost all map sets. The average collision times of the solution by CBS-MPC are between 7 to 10 under different map sizes, which are all considered as failures. The agents would collide with each other when using MPC following their guided paths since the kinematic constraints are not considered in the high-level CBS. Our proposed method, however, has no collisions in any of the map sets and performs smaller makespans in 300$\times$300m and 100$\times$100m map sets as well.

\subsubsection{Comparision with HA*}
We assess then how our method scales to a large number of robots compared to the decentralized method HA*, seeing that CBS-MPC have collisions between agents in most cases. The sequential version of our method (with $K_b$=2) also participates in the comparison. We limit the computation time of each instance to 120 seconds, and the results are shown in Fig.\ref{Scalability result}.

In 300$\times$300m map sets, our CL-CBS approach outperforms HA* in both success rate and runtime. Our original method successfully solves over 50\% of instances containing 60 agents in both empty and obstructive scenarios and the sequential version solves instances including 100 agents with over 70\% success rate.  On the contrary, HA* barely works out instances over 50 agents.
The running time of HA* is more than twice as long as ours in the same map set. As for 100$\times$100m map sets, our method solves instances up to 40 agents in 30 seconds while HA* costs over 60 seconds for instances containing 30 agents with success rate below 10\%.

\subsubsection{Comparision with sequential version}
As we proposed the sequential version of our method in \ref{Sequential Car-like CBS}, we evaluate the performance of the original method and its sequential version with different batch numbers $K_b$. We perform a comparative test in two map sets: a large map with obstacles (300x300\_agents60\_obs) and a small empty one (50x50\_agents15\_empty). The time limit for each instance is 60 seconds. The results are shown in Fig.\ref{Sequential result}.

In the large map set, our original method ($K_b$=1) achieves merely 38.3\% success rate and costs 31.6 seconds runtime at average.  Fig.\ref{Sequential1} shows the success rate increases rapidly to 98.3\%, and the average runtime decreases to 6.8 seconds when we divide agents into three batches. The runtime reduced to 4.4 seconds when $K_b$=6, which is almost an order of magnitude smaller than the original method. Meanwhile, the average sum of costs has only increased by 2.5\%.

As we mentioned before, the sequential method sacrifices completeness guarantee and may lead to some fail cases. In the 50$\times$50m map set, when $K_b$ increases, the success rate first rises to 98.3\% and then falls to 86.6\% as Fig.\ref{Sequential2}.  This is for the reason that previous agents are not aware of the existence of subsequent agents during the planning and may block their paths to the goal.

\subsection{Field Test}
We conduct field tests using seven 23$\times$20 cm Ackermann-steering robots produced by WeTech as shown in Fig.\ref{fig1}. The robot is able to move at 0.3m/s, and the minimum turning radius is 26cm. All the robots are equipped with a 2D Lidar from Slamtec, a 5-megapixel camera, and a Raspberry Pi 4 running Ubuntu 18.04 and ROS Melodic. We use a PC laptop running ROS as the central computing station to communicate with all agents use 2.4GHz Wifi.

Experiments are performed in a 5$\times$3m room, including empty and obstructed scenarios. Before each experiment, we create a 2D occupancy map using lidar by gmapping algorithm. After appointing start and goal states for all agents, a solution is computed on the laptop. We then transfer paths to a sequence of velocity commands and send them to agents for execution. Amcl package is used when robots are running so that we can get the trajectory. The snapshot of the field test is shown in Fig.\ref{experiment snapshot}, and full experiments are presented in the supplemental video.

\section{Conclusion and Futur Works}
\label{Conclusion and Futur Works}

In this paper, we formalize the CL-MAPF problem that considers the kinematic and spatiotemporal  constraints of car-like robots. We present CL-CBS, an efficient hierarchical search-based algorithm that is correct and complete for solving the problem. Experiments in simulated and physical environments show that our method outperforms baseline solvers in terms of both scalability and solution quality.
One of the directions of future research is extending our method in order to plan for holonomic and nonholonomic agents under the same scenario, and another one is applying the proposed approach to combined target-assignment and path-finding (TAPF) problem.


\addtolength{\textheight}{-4cm}

\bibliographystyle{IEEEtran}
\bibliography{cite.bib}

\end{document}